\documentclass[letterpaper, 10 pt, conference]{ieeeconf}  

\IEEEoverridecommandlockouts                              

\usepackage{cite}
\usepackage{amsmath,amssymb,amsfonts}
\usepackage{algorithmic}
\usepackage{graphicx}
\usepackage{tabularx}
\usepackage{textcomp}
\usepackage{cuted}
\usepackage{capt-of}
\usepackage{float}
\usepackage{booktabs}
\usepackage{threeparttable}
\usepackage{mathtools}
\usepackage[mathscr]{eucal}
\usepackage[table,xcdraw]{xcolor}
\usepackage{colortbl}
\usepackage{colortbl}
\usepackage{balance}

\usepackage{marvosym}
\usepackage{algorithmic}
\usepackage{indentfirst}
\usepackage{authblk}

\makeatletter
\let\NAT@parse\undefined
\makeatother
\usepackage{hyperref}

\title{PUGS: Zero-shot \underline{P}hysical \underline{U}nderstanding with \underline{G}aussian \underline{S}platting
}

\author{Yinghao Shuai$^{1,8}$, Ran Yu$^{2}$, Yuantao Chen$^{3,8}$, Zijian Jiang$^{1,8}$, Xiaowei Song$^{1,8}$, Nan Wang$^{1,8}$, \\ Jv Zheng$^{6,8}$, Jianzhu Ma$^{6}$, Meng Yang$^{4,5}$, Zhicheng Wang$^{1}$, Wenbo Ding$^{2}$, and Hao Zhao$^{\dag,6,7,8}$
}

\begin{document}


\twocolumn[{%
  \renewcommand\twocolumn[1][]{#1}%
  \maketitle
  \centering
  \includegraphics[width=0.99\textwidth]{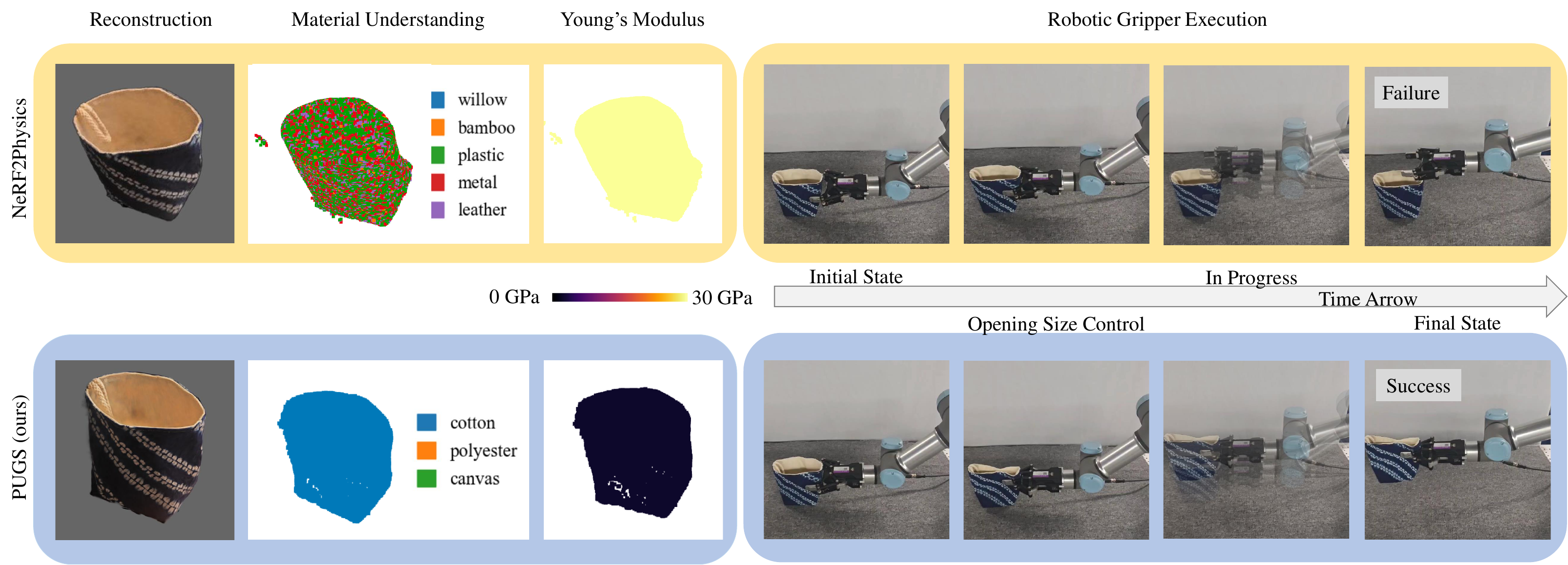}
  
  \captionof{figure}{\textbf{Comparison between NeRF2Physics \cite{zhai2024physical} and our proposed PUGS.} The input is a set of RGB images, and the output is a reconstructed target with physical property understanding. PUGS switches from NeRF to Gaussian Splatting and correctly predicts both the material (cotton) and the Young's modulus, while NeRF2Physics fails. With the correct Young's modulus, the robotic gripper can adjust its opening size and successfully grasp the object.}
  \label{fig:teaser}
}]

\renewcommand{\thefootnote}{}
    \footnotetext{\dag Corresponding author.}
\renewcommand{\thefootnote}{\arabic{footnote}} 
\footnotetext[1]{School of Computer Science and Technology, Tongji University, $\ $China, \{shuaiyinghao,jiangzj,kevin729,2233017,zhichengwang\}@tongji.edu.cn}
\footnotetext[2]{Shenzhen Ubiquitous Data Enabling Key Lab, Shenzhen International Graduate School, Tsinghua University, Shenzhen 518055, China,  yur23@mails.tsinghua.edu.cn and ding.wenbo@sz.tsinghua.edu.cn}
\footnotetext[3]{The Chinese University of Hong Kong, Shenzhen, China, 224015065@link.cuhk.edu.cn}
\footnotetext[4]{MGI Tech, Shenzhen 518083, China}
\footnotetext[5]{Chulalongkorn University, 10330, Bangkok, Thailand}
\footnotetext[6]{Institute for AI Industry Research (AIR), Tsinghua University, China, zhaohao@air.tsinghua.edu.cn}
\footnotetext[7]{Beijing Academy of Artificial Intelligence(BAAI), China}
\footnotetext[8]{Lightwheel AI}

\thispagestyle{empty}
\pagestyle{empty}

\begin{abstract}
Current robotic systems can understand the categories and poses of objects well. But understanding physical properties like mass, friction, and hardness, in the wild, remains challenging. We propose a new method that reconstructs 3D objects using the Gaussian splatting representation and predicts various physical properties in a zero-shot manner. We propose two techniques during the reconstruction phase: a geometry-aware regularization loss function to improve the shape quality and a region-aware feature contrastive loss function to promote region affinity. Two other new techniques are designed during inference: a feature-based property propagation module and a volume integration module tailored for the Gaussian representation. Our framework is named as zero-shot physical understanding with Gaussian splatting, or PUGS. PUGS achieves new state-of-the-art results on the standard benchmark of ABO-500 mass prediction. We provide extensive quantitative ablations and qualitative visualization to demonstrate the mechanism of our designs. We show the proposed methodology can help address challenging real-world grasping tasks. Our codes, data, and models are available at https://github.com/EverNorif/PUGS
\end{abstract}

\section{Introduction}

The current robotic vision systems have made significant strides in recognizing object categories \cite{cheng2024yolo}\cite{chen2022pq} and poses \cite{wen2024foundationpose}\cite{zhong2020seeing}. However, for effective grasping and manipulation in unstructured environments, it is crucial to also understand key physical properties of objects. For instance, as shown in Fig.~\ref{fig:teaser}, a cotton bag is soft, and if the planning algorithm selects a large gripper opening size, the bag could slip, resulting in a failed grasp. Similarly, understanding an object’s mass and friction is essential for robotic grippers to handle heavy or slippery items. Therefore, this paper focuses on reconstructing objects while inferring their physical properties in a zero-shot fashion.

We build on the framework of NeRF2Physics \cite{zhai2024physical}, which takes a set of RGB images as input, reconstructs the target object, and densely predicts its physical properties. Drawing inspiration from the recent success of 3D Gaussian Splatting (3DGS) \cite{kerbl3Dgaussians}, known for its efficiency and high-quality rendering, we incorporate 3DGS into this task and develop the PUGS framework—standing for zero-shot \textbf{P}hysical \textbf{U}nderstanding with \textbf{G}aussian \textbf{S}platting. We identify two key limitations of the vanilla 3DGS: (1) A lack of geometric regularization, where floating Gaussian primitives may blend effectively in RGB volume rendering but introduce difficulties in volume integration for physical properties like mass. (2) A lack of local affinity, as demonstrated in the top-left panel of Fig.~\ref{fig:teaser}, where material understanding results are fragmented, making them impractical for downstream tasks such as grasping and manipulation planning.

To address these two issues, we introduce a geometry-aware regularization loss and a region-aware feature contrastive loss. The geometry-aware regularization loss ensures that the 3D Gaussians align more closely with the object’s spatial structure. The region-aware feature contrastive loss guides the feature vectors of the reconstructed Gaussians to reflect the local region affinity observed in RGB images.

Building on the reconstruction results, we integrate the Vision-Language Foundation Model (VLM) to perform zero-shot physical property prediction. These predictions are propagated to all Gaussians using region-aware features, resulting in dense reconstructions with physical properties. Given the importance of object-level physical properties, such as mass, we propose a Gaussian-based volume integration module. We evaluate PUGS on the mass prediction benchmark introduced by NeRF2Physics\cite{zhai2024physical} and achieve new state-of-the-art (SOTA) results. Visualizations further demonstrate that PUGS provides accurate material understanding and reliable physical property predictions.

In brief, our contributions can be summed up in three-fold:
\begin{itemize}
\item[$\bullet$] We present PUGS, which exploits 3D Gaussian Splatting for efficient object reconstruction and zero-shot physical property inference from RGB images.
\item[$\bullet$] To address limitations in 3DGS, we propose geometry-aware regularization and region-aware feature contrastive losses, improving alignment with object geometry and capturing local region affinities.
\item[$\bullet$] We incorporate Vision-Language Models for zero-shot physical property prediction and introduce a Gaussian-based volume integration module for object-level properties like mass, achieving state-of-the-art results.
\end{itemize}

\section{Related Works}

\subsection{3D Reconstruction}
Recently, Neural Radiance Field (NeRF)\cite{mildenhall2020nerf} has raised a lot of attention due to its photo-realistic rendering ability. 3D Gaussian Splatting (3DGS)\cite{kerbl3Dgaussians} goes one step further to achieve real-time rendering. Some works have extended NeRF\cite{mildenhall2020nerf} and 3DGS\cite{kerbl3Dgaussians} to address their inherent limitations\cite{chen2023nerrf3dreconstructionview, Yuan_2024, liu2024ripnerf, song2024sa} or leverage these novel 3D representations for applications in other domains\cite{wu2023mars, peng2024synctalk, zhang2024droneassistedroadgaussiansplatting}. However, most of these works focus on delivering better visual quality but overlook the importance of physical properties, which is critical for application in robotics.
Some previous works focus on the accurate geometry\cite{guedon2023sugar, Huang2DGS2024, zhang2024radegsrasterizingdepthgaussian, chen2024pgsr}, in which PGSR\cite{chen2024pgsr} adds a new geometry loss that promises normal-depth consistency. NeRF2Physics\cite{zhai2024physical} utilizes NeRF\cite{mildenhall2020nerf} as a 3D representation and proposes to predict its physical property using Large Language Model (LLM) for the first time. Yet it tries to extract point clouds from NeRF as the scene's geometry to calculate mass, which is slow and may not be accurate enough.

\subsection{Physical Property Prediction}
Physical property prediction from visual data is very important for robotics\cite{zhang2024adaptigraph, shi2023robocook, li2018learning}. However, collecting paired data between images and various physical properties is very hard. Some previous works propose reasoning physical properties by observing the object's movement or interaction with other objects in a 3D physical engine\cite{li2023pacnerf, 10.1007/978-3-319-46475-6_1, NIPS2015_d09bf415, 10160731}. However, these methods are still limited to a few physical properties and are hard to use. NeRF2Physics\cite{zhai2024physical} is the first to utilize LLMs for physical property prediction tasks in a zero-shot manner and Octopi\cite{yu2024octopi} further proves its importance in grasping tasks. We aim to make the process faster and more accurate by using 3DGS and a more reliable framework.

\subsection{Vision Language Model}
Vision Language Models (VLMs)\cite{ilharco_gabriel_2021_5143773,zhu2023minigpt,li2023blip} have become increasingly popular in robotics, enabling applications across various tasks\cite{huang2023voxposer,moo2023arxiv,chen23polarnet,rt22023arxiv,qin2024langsplat,zhou2024feature,kerr2023lerf,zhou2024navgpt}. While some methods\cite{zhou2024navgpt,huang2023voxposer} utilized text and images as input to plan complex tasks, others, like\cite{kerr2023lerf,qin2024langsplat}, employ CLIP\cite{ilharco_gabriel_2021_5143773} to establish a feature distillation field for aligning 3D representions with text. In our work, We fully leverage the capabilities of VLMs. By utilizing GPT-4 for single-image physical property prediction and CLIP\cite{ilharco_gabriel_2021_5143773} for mapping these properties to 3D reconstructions, we achieve zero-shot physical property prediction with 3D reconstruction.

\section{Method}\label{sec:method}

\subsection{Overview}\label{sec:method-overview}

\begin{figure*}
\vspace{2pt}
  \centering
  \includegraphics[width=0.95\textwidth]{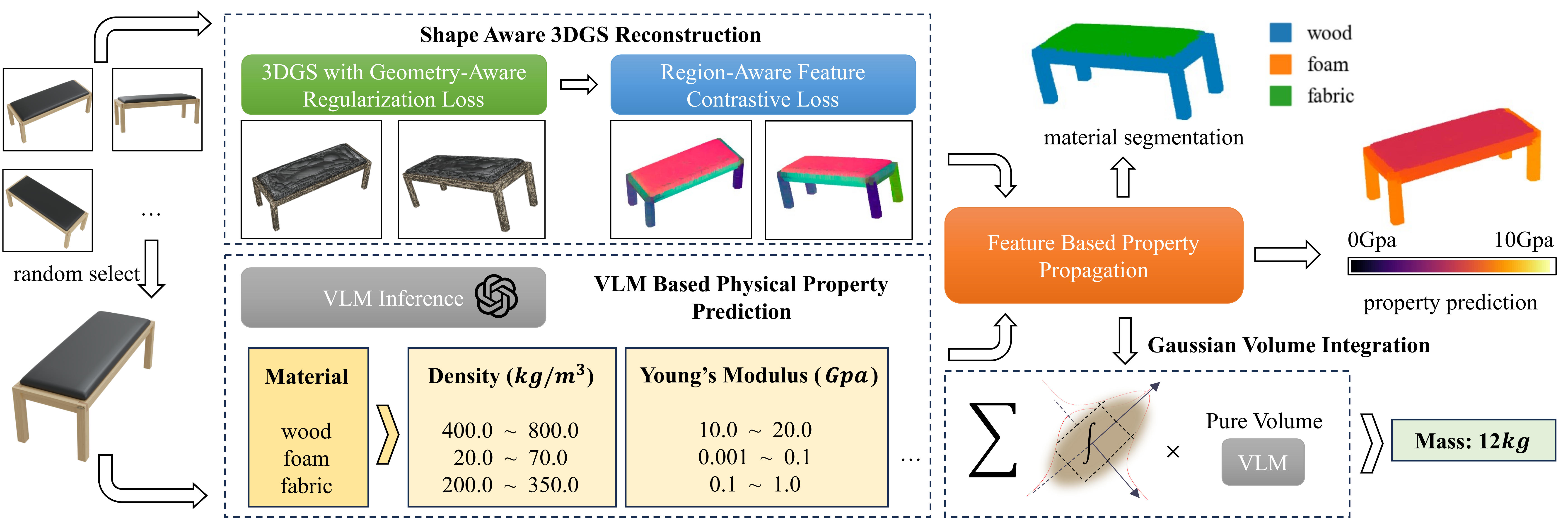}
  \caption{\textbf{Overview of PUGS}. We take the multi-view images of the object as input, reconstruct the Gaussians representation with regional features through Shape Aware 3DGS Reconstruction, predict the candidate properties in a zero-shot manner through VLM Based Physical Property Prediction, and finally obtains the material segmentation result and property prediction result through Feature Based Property Propagation. With proposed Gaussian Volume Integration, we can calculate the object-level property, like mass.}
  \label{fig:pipeline}
\end{figure*}

As shown in the Fig.~\ref{fig:pipeline}, our proposed PUGS can reconstruct the object from the calibrated multi-view images $\mathcal{I}$ and subsequently predict the multiple physical properties in a zero shot manner. PUGS can be divided into three phase: shape aware 3DGS reconstruction (Sec.~\ref{sec:method-3dgs}), VLM based physical property prediction (Sec.~\ref{sec:method-vlm}) and feature based property propagation (Sec.~\ref{sec:method-propagation}). Firstly, we reconstruct the object using 3DGS, with the geometry-aware regularization loss employed to ensure consistency between the Gaussian and the actual spatial shape distribution of the object. At the same time, we utilize SAM\cite{kirillov2023segany} and contrast learning to train region-aware features. Subsequently, the common-sense capabilities and prior knowledge of VLM are employed to predict the physical properties, which are then distilled into the reconstruction results through the combination of CLIP features. Finally, we introduce a way to compute object-level properties based on Gaussian volume integration (Sec.~\ref{sec:gs_integraton}).

\subsection{Shape Aware 3DGS Reconstruction}\label{sec:method-3dgs}

\textbf{Preliminary of 3D Gaussian Splatting.} 3DGS\cite{kerbl3Dgaussians} represents a novel approach for scene representation that not only delivers high-quality rendering in a relatively short training period but also supports real-time rendering and novel view synthesis. Given calibrated multi-view images $\mathcal{I}$, 3DGS explicitly reconstructs a scene with a realistic rendering result utilizing a set of 3D Gaussians $\{\mathcal{G}_i\}$. Each Gaussian can be represented as follows:
\begin{equation}
    \mathcal{G}_i(\mathbf{x}|\mathbf{\mu}_i,\mathbf{\Sigma}_i) = e^{-\frac{1}{2}(\mathbf{x}-\mathbf{\mu}_i)^T\mathbf{\Sigma}_i^{-1}(\mathbf{x}-\mathbf{\mu}_i)}
\end{equation}
where $\mathbf{\mu}_i \in \mathbb{R}^3$ is the center position of each Gaussian and $\mathbf{\Sigma}_i \in \mathbb{R}^{3\times 3}$ is the corresponding 3D covariance matrix, which can be decomposed into a combination of rotation matrix $\mathbf{R}$ and scaling matrix $\mathbf{S}$.
\begin{equation}
    \mathbf{\Sigma}_i = \mathbf{R}_i\mathbf{S}_i\mathbf{S}_i^T\mathbf{R}_i^T
\end{equation}
Additionally, each 3D Gaussian has an opacity $\sigma$ and spherical harmonic coefficient SH, which is used to characterise viewpoint-dependent colors.

3DGS uses fast $\alpha$-blending for rendering. Given a transformation $W$, an intrinsic $K$ and Jacobian matrix $J$, $\mathbf{\mu}_i$ and $\mathbf{\Sigma}_i$ can be transformed to camera coordinate and then projected to NDC space. $\alpha$-blending use the following formula to get the color $C$ of each pixel:
\begin{equation}\label{eq:a-blending}
    C = \sum_{i\in N} T_i \alpha_i c_i, \quad T_i = \prod_{j=1}^{i-1}(1-\alpha_j)
\end{equation}
where $\alpha_i$ is calculated with the related learnable opacity $\sigma$, $c_i$ is the view-dependent color.

With the rendered images, the loss function is the $\mathcal{L}_1$ loss and SSIM loss with the GT images:
\begin{equation}
    \mathcal{L}_{\rm 3dgs} = (1-\lambda)\mathcal{L}_1+\lambda \mathcal{L}_{\rm SSIM}
\end{equation}

\textbf{Geometry-Aware Regularization Loss.} The vanilla 3DGS only uses RGB loss as supervision, which can easily lead to local optima. This results in the distribution of 3D Gaussian being less compatible with the actual shape of the object, which ultimately leads to poor geometry. Our goal is to reconstruct the object with physical understanding and serve it in the downstream tasks of simulation environment, which impose higher demands on the geometry of 3D Gaussian. To address this challenge, we introduce the geometry loss in \cite{chen2024pgsr}:
\begin{equation}
    \mathcal{L}_{\rm geo} = \sum_{I \in \mathcal{I}}\frac{1}{W(I)}\sum_{p \in W(I)} |\bar{\triangledown I}|^5 ||\mathbf{N}_d(p) - \mathbf{N}(p)||_1
\end{equation}
where $W(I)$ is the set of pixel in image $I$, $\bar{\triangledown I}$ is the image gradient normalized to the range of 0 to 1, $\mathbf{N}_d(p)$ is calculated by local plane of the pixel point $p$ and $\mathbf{N}(p)$ is calculated with unbiased depth rendering\cite{chen2024pgsr}. Geometry loss can maintain consistency between depth and normal geometry, providing fairly accurate geometric information.

Additionally, we use sparse loss\cite{xu2022point} to encourage Gaussian sphere's opacity value $\sigma$ to approach either 0 or 1, which helps encourage the distribution of 3D Gaussian to be more consistent with the actual spatial shape of the object:
\begin{equation}
    \mathcal{L}_{\rm sparse} = \frac{1}{|\sigma|}\sum_{\sigma}[\log (\sigma_i) + \log (1-\sigma_i)]
\end{equation}

Finally, the total loss of Gaussian training includes image loss and geometry-aware regularization loss composed of $\mathcal{L}_{\rm geo}$ and $\mathcal{L}_{\rm sparse}$:
\begin{equation}
    \mathcal{L} = \mathcal{L}_{\rm 3dgs} + \lambda_1 \mathcal{L}_{\rm geo} + \lambda_2 \mathcal{L}_{\rm sparse}
\end{equation}

\begin{figure}
  \centering
  \includegraphics[width=0.48\textwidth]{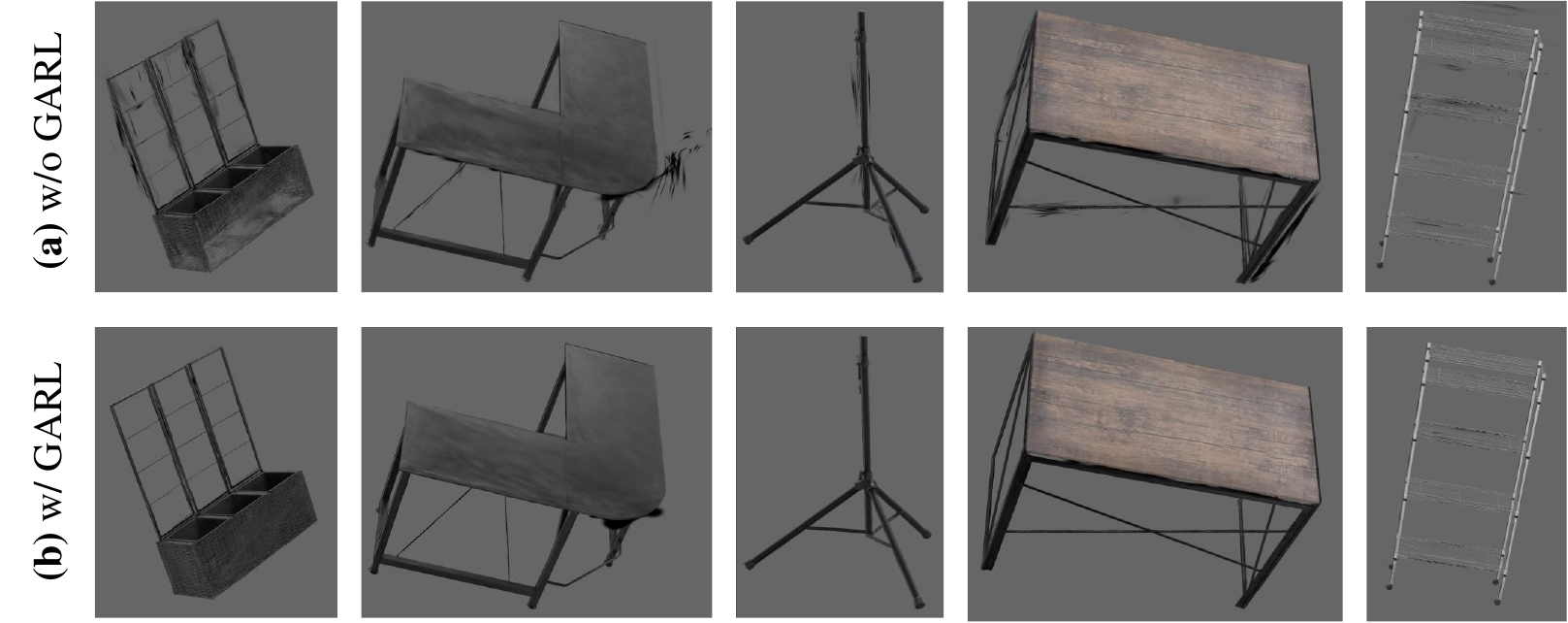}
  \caption{\textbf{Comparison of reconstruction results before and after applying geometry-aware regularization loss (GARL).} Result (a) without the GARL exhibit some floaters and blurred areas; result (b) with the GARL can achieve the results with more accurate geometry. }
  \label{fig:shape-geometry-loss}
\end{figure}

\begin{figure*}
\vspace{3pt}
  \centering
  \includegraphics[width=0.95\textwidth]{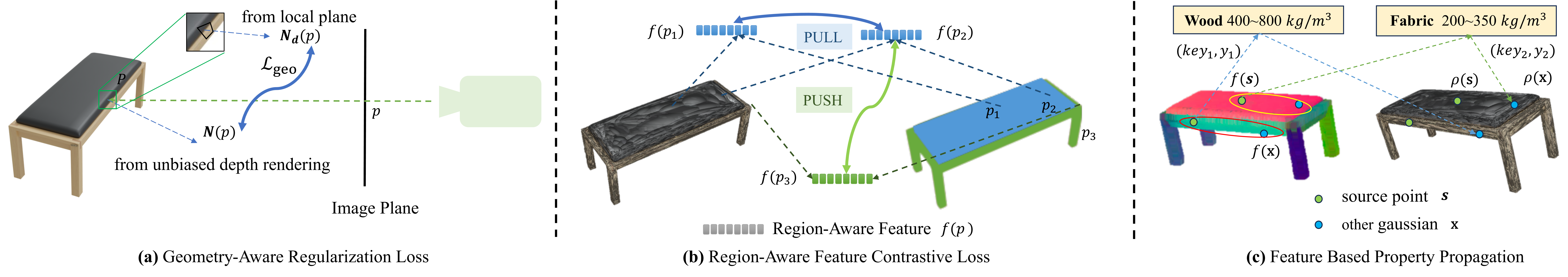}
  \caption{\textbf{Explanation of different modules in PUGS}. During the reconstruction process, we compute the (a) geometry-aware regularization loss using normals obtained through two different methods. With (b) region-aware feature contrastive loss, we pull the features corresponding to Gaussians belonging to the same mask, while pushing apart the features corresponding to different masks. During (c) feature based property propagation, we use the  similarity of region-aware feature to propagate physical properties.}
  \label{fig:method-explain}
\end{figure*}

\textbf{Region-Aware Feature Contrastive Loss.} An object be usually composed of a variety of materials, which are typically associated with specific regions. As illustrated in the Fig.~\ref{fig:compare-result}(a), different regions represent distinct materials. The efficacy of material prediction can be enhanced by accurately identifying the different regions of an object in the reconstruction results, and then lead to an improvement of physical property prediction. In order to achieve this, we follow SAGA\cite{cen2023saga} by refining the segmentation capability of SAM\cite{kirillov2023segany} from 2D Masks to additional features in Gaussian. 

Given the calibrated multi-view images, we firstly reconstruct an initial Gaussian representation using RGB loss and geometry-aware regularization loss. Subsequently, we introduce a region-aware feature $f \in \mathbb{R}^{D}$ for each Gaussian, where $D$ is the dimension of feature. Following the similar approach of $\alpha$-blending (Eq.~\ref{eq:a-blending}), the features can be rendered into a 2D feature map $\mathbf{F}$.

We utilize contrastive learning to train this feature, enabling it to develop region-aware capabilities. We use SAM to extract the mask maps from multi-view images.  As illustrated in
the Fig. ~\ref{fig:method-explain}(b), if two pixels belong to the same mask, the corresponding features should have a higher similarity. Therefore, the features can be trained with the following contrastive learning loss:
\begin{align}
    \mathcal{L}_{\rm coor}(p_1, p_2) &= [1 - 2\text{coor}(p_1, p_2)]\max[\text{coor}_{\rm f}(p_1, p_2), 0], \nonumber \\
    \text{coor}_{\rm f}(p_1, p_2) &= <\mathbf{F}(p_1), \mathbf{F}(p_2)>
\end{align}
where $\text{coor}(p_1, p_2)$ indicates whether the pixel $p_1$ and $p_2$ belong to the same mask, and $\text{coor}_{\rm f}(p_1, p_2)$ is defined as the cosine similarity between corresponding features of $p_1$ and $p_2$. Following the SAGA implementation, we also use the feature norm regularization, and the final loss of the region-aware feature in one training image $I$ is:
\begin{align}
    \mathcal{L}_{\rm region} = \sum_{(p_1,p_2)\in W(I)\times W(I)} \mathcal{L}_{\rm coor}(p_1,p_2) \nonumber\\
    + \frac{1}{HW}\sum_{p\in W(I)} (1 - ||\mathbf{F}(p)||_2)
\end{align}

In this stage, we only train the region-aware feature on Gaussian, without optimizing other parameters. After this stage of self-supervised training, the region-aware feature on Gaussian is capable of representing regions of objects. If two Gaussian represent the same region, their corresponding feature will demonstrate a high level of similarity.

\subsection{VLM Based Physical Property Prediction}\label{sec:method-vlm}

After reconstruction stage in Sec.~\ref{sec:method-3dgs}, the object can be reconstructed using 3D Gaussian, and regionfeature can help to distinguish different regions. Subsequently, we make predictions regarding the physical properties based on the reconstruction results, including its density, hardness, Young's modulus, and so on.

Similarly to NeRF2Physics\cite{zhai2024physical}, we believe that large model exhibits robust common-sense capabilities that can facilitate zero-shot physical understanding. There are some differences in the specifics between our proposed PUGS and NeRF2Physics,.

NeRF2Physics employs a two-stage methodology to achieve the prediction of material and physical properties. Initially, the VQA model (e.g., BLIP2\cite{li2023blip}) is employed to obtain a description of the image. Subsequently, the image description is feed into the large language model, which is then used to predict the material and physical properties. We observe that a certain degree of information is lost during the image-to-text process. To illustrate, we can find the wrong result about a baskets in the rightmost column of Fig.~\ref{fig:compare-result}. The VQA model predicts the description is three black baskets with a black handle, but the large language model considers it to be the baskets made of rattan, which results in incorrect material understanding.

In contrast to NeRF2Physics, we employ the VLM directly for the physical understanding. One of the multi-view images $\mathcal{I}$ is randomly selected as the input to the VLM, which is then prompted to perform the image description and the prediction of material and physical properties. Finally, we can get a dictionary of $K$ candidate materials $\mathcal{M} = \{(key_k, y_k)\}$, where $key_k$ is the material name and $y_k$ is the specific physical property value or value range. 

\subsection{Feature Based Property Propagation}\label{sec:method-propagation}
In order to integrate the predicted material and physical properties into the reconstruction result, we employ CLIP\cite{radford2021learningtransferablevisualmodels} as a bridge between disparate modalities. By considering the reconstructed 3D Gaussian as point cloud and performing voxel down-sampling, a set of source points $\mathcal{S}$ distributed on the surface of the object can be obtained. For each source point, we project it onto each input view according to the corresponding camera parameters, and use depth test to determine the occlusion. If the source point is visible on image $I$, the projection point on the image plane is identified as the centre, and a $p \times p$ patch is extracted to get CLIP feature vector. Ultimately, the CLIP feature $\mathbf{z}$ of each source point will be the average of the CLIP features on different images. With the CLIP feature of the source point, the predicted material and physical properties are fused to the corresponding Gaussian. This is achieved by the following formula:
\begin{equation}
    \rho(\mathbf{s}) = \frac{\sum_{k=1}^K\exp(\omega_k[\mathbf{s}]/T)y_k}{\sum_{k=1}^K\exp(\omega_k[\mathbf{s}]/T)}
\end{equation}
where $\rho(\mathbf{s})$ is the physical property value of source point $\mathbf{s}$, $\omega_k[\mathbf{s}] = \phi_{CLIP}(\mathbf{z}, key_k)$ is the cosine similarity between CLIP feature of source point $\mathbf{s}$ and the language CLIP feature of material name $key_k$, $T$ is a temperature parameter. 

\begin{figure*}
  \centering
  \includegraphics[width=0.99\textwidth]{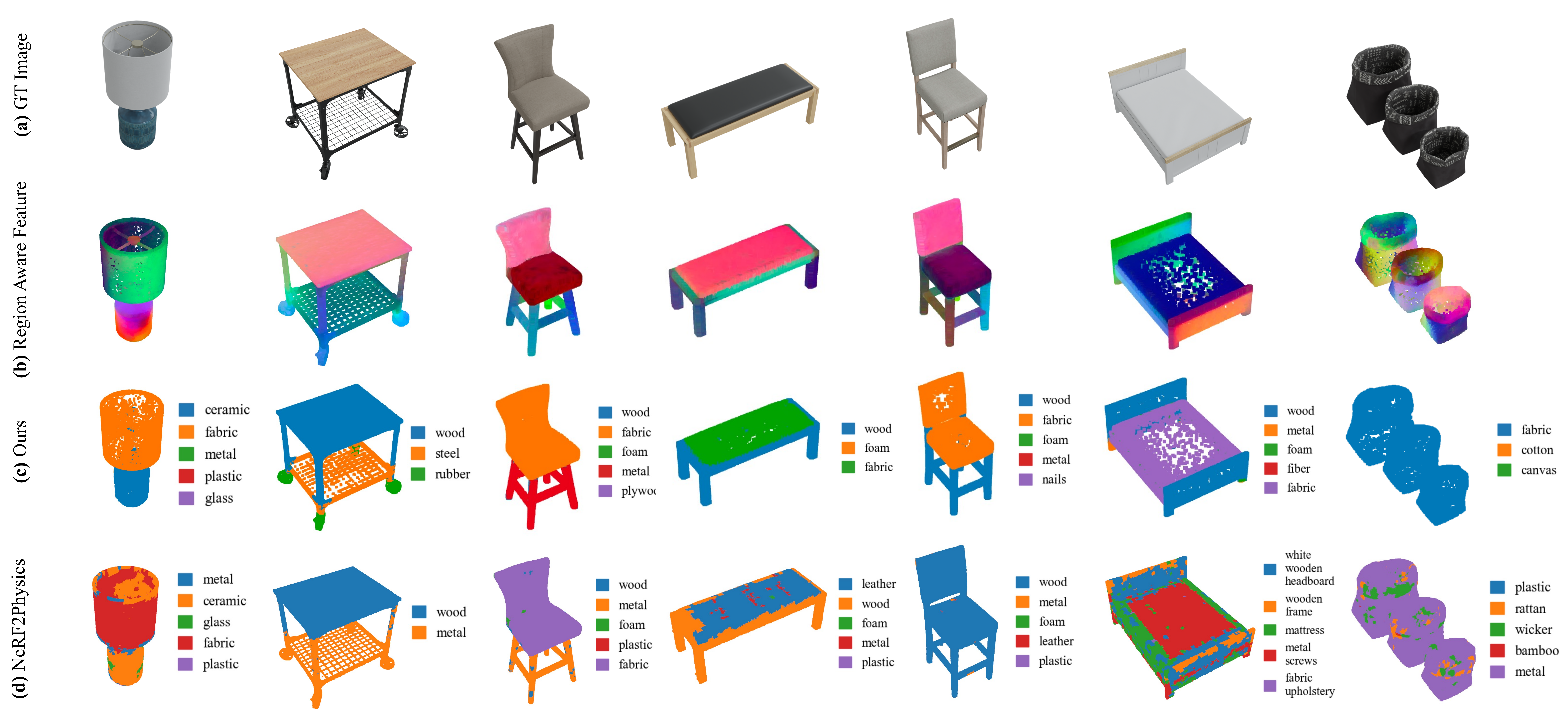}
  \caption{\textbf{Region Aware Feature Visualization and Material Segmentation Result of some object.} The visualization results in (b) demonstrate that our region-aware feature can effectively identify different regions of the object. (c) and (d) represent the material prediction results of PUGS and NeRF2Physics, respectively. The results from NeRF2Physics are more fragmented, whereas our PUGS achieves more coherent and accurate material segmentation.}
  \label{fig:compare-result}
\end{figure*}

After that, we employ interpolation to propagate the physical property from the source point. NeRF2Physics\cite{zhai2024physical} utilizes nearest-neighbor interpolation. However, this approach may result in a more dispersed and fragmented outcome for the material segmentation, as illustrated in Fig.~\ref{fig:compare-result}(d). In contrast, PUGS employs a interpolation based on the similarity of region-aware feature as illustrated in the Fig.~\ref{fig:method-explain}(c):
\begin{equation}
    \rho(\mathbf{x}) = \rho(\text{argmax}_{\mathbf{s}\in \mathcal{S}} <f(\mathbf{x}), f(\mathbf{s})>)
\end{equation}
where $f(\mathbf{x})$ and $f(\mathbf{s})$ are the region-aware feature of Gaussian $\mathbf{x}$ and $\mathbf{s}$ respectively. As shown in Fig.~\ref{fig:compare-result}(c), our interpolation based on region-aware feature enables the attainment of more uniform and precise material segmentation results.

\subsection{Gaussian Volume Integration}\label{sec:gs_integraton}

Following NeRF2Physics\cite{zhai2024physical}, we also consider the prediction of object-level physical properties, such as mass, which requires the estimation and integration of the volume of an object. In contrast to the approach taken in NeRF2Physics, where volume estimation is based on the predicted material thickness, we employ the explicit properties of 3D Gaussian for the integration. Each 3D Gaussian is a 3D ellipsoid in the space, and the corresponding opacity $\sigma$ can be used to weight it. The associated point-level physical properties are then multiplied by the weight, and perform a cumulative computation to obtain the object-level physical properties. 

Benefit from the geometry loss in Sec.~\ref{sec:method-3dgs}, the reconstructed 3D Gaussian is distributed along the surface of the object. However, the reconstructed objects with 3D Gaussian will exhibit substantial internal voids, so the estimated result is the volume of the surface layer of the object, which still differs from the actual volume. In order to compensate for this discrepancy, we introduce the concept of pure volume. The pure volume of an object is defined as the volume it occupies after the removal of all hollow areas. Consequently, the pure volume of an object is typically smaller than the spatial volume that is commonly perceived. Only when an object is completely solid, its spatial volume equals to the pure volume. Furthermore, the concept of pure volume provides the insight into the scale of the object, which enhances the precision of our predictions regarding object-level property.

We prompt the VLM to get the pure volume $v$ of one object. The specific prompts can be referenced from the implementation details. Then we use it to revise our prediction and get the final prediction result $m$:
\begin{equation}
    m = \frac{v}{c}\hat{m}
\end{equation}
where $\hat{m}$ is the predicted object-level physical property in the above step, $c$ is the predicted volume of all 3D Gaussian.

\section{Experiments}
Firstly, we evaluate PUGS on the ABO-500 dataset, which serves as the benchmark for object mass estimation in NeRF2Physics\cite{zhai2024physical}. Subsequently, PUGS is adapted for downstream applications, enabling the robotic arm to perform object grasping based on the predicted physical property.



\subsection{Object Mass Estimation}

\textbf{Dataset.} We utilize the ABO-500 dataset for object mass estimation. The ABO-500 is a subset derived from the Amazon Berkeley Objects (ABO) dataset\cite{collins2022abo}, initially used by NeRF2Physics. The ABO-500 subset specifically includes 500 objects from ABO dataset, each accompanied by calibrated multi-view images and object mass annotations.

\textbf{Metrics.} Follow NeRF2Physics, we also use the same metrics for evaluation, including Absolute Difference Error (ADE), Absolute Log Difference Error (ALDE), Absolute Percentage Error (APE) and Min Ratio Error (MnRE).

\textbf{Qualitative Results.} Fig.~\ref{fig:compare-result} presents a comparison of material segmentation results between our PUGS and NeRF2Physics. It can be observed that our pipeline produces more coherent and reliable material segmentation with fewer fragmented results.

\textbf{Quantitative Results.} We report the mass estimation metrics on the whole ABO-500 in Tab.~\ref{tab:result}, where bold number means better result. The result shows that our pipeline achieved superior results across all evaluated metrics.

\textbf{Ablations.} We perform ablations with different approach settings. Firstly, we remove the geometry-aware regularization loss. The results indicate that geometry-aware regularization loss provides a modest improvement. Fig.~\ref{fig:shape-geometry-loss} shows that geometry-aware regularization loss enhances the geometric structure of objects with fine details. Therefore, we select 100 objects (subset A) from the ABO-500 dataset that possess more intricate structures, such as shelves with mesh designs. As shown in the Tab.~\ref{tab:ablation-result-sgl}, a greater average improvement can be observed on these objects. Next, we remove the region-aware feature training. Since the majority of objects are dominated by a single material, removing it results in only a slight performance drop. However, as shown in the Fig.~\ref{fig:compare-result}, it helps produce more reliable material segmentation. We also select 100 objects (subset B) composed of multiple materials. As indicated in the Tab.~\ref{tab:ablation-result-raft}, region-aware feature leads to a more significant improvement for these objects. Lastly, we change our Gaussian volume integration to thickness based integration from NeRF2Physics\cite{zhai2024physical}. The results indicate that the thickness-based volume integration method is not suitable for our Gaussian representations. It is difficult to find a single appropriate correction factor for all objects.

\begin{table}[t]
\centering
\caption{Quantitative results of object mass estimation on ABO-500.}
\resizebox{0.48\textwidth}{!}{
\begin{tabular}{@{}llcccc@{}}
\toprule
Approach     &  & \multicolumn{1}{l}{ADE ($\downarrow$)} & \multicolumn{1}{l}{ALDE ($\downarrow$)} & \multicolumn{1}{l}{APE ($\downarrow$)} & \multicolumn{1}{l}{MnRE ($\uparrow$)} \\ \midrule
NeRF2Physics &  & 12.725  & 0.736 & 1.040 & 0.564 \\
PUGS (Ours) &  & \textbf{9.461} & \textbf{0.661} & \textbf{0.767} & \textbf{0.576}  \\ 
 &  & \textcolor{red}{+25.65\%} & \textcolor{red}{+10.19\%} & \textcolor{red}{+26.25\%} & \textcolor{red}{+2.13\%}  \\ 
\bottomrule
\end{tabular}
}
\label{tab:result}
\end{table}

\begin{table}[t]
\centering
\caption{Ablation about geometry-aware regularization loss (\textbf{GARL}).}
\resizebox{0.48\textwidth}{!}{
\begin{tabular}{@{}ccccc@{}}
\toprule
Approach Setting & \multicolumn{1}{l}{ADE ($\downarrow$)} & \multicolumn{1}{l}{ALDE ($\downarrow$)} & \multicolumn{1}{l}{APE ($\downarrow$)} & \multicolumn{1}{l}{MnRE ($\uparrow$)} \\ \midrule
w/o GARL (500 objects) & 9.804 & 0.675 & 0.858 & 0.567 \\
w/ GARL (500 objects) & \textbf{9.461} & \textbf{0.661} & \textbf{0.767} & \textbf{0.576} \\ 
&  \textcolor{red}{+3.50\%} & \textcolor{red}{+2.07\%}& \textcolor{red}{+10.60\%}& \textcolor{red}{+1.59\%}\\
\midrule
w/o GARL (subset A) & 5.332 & 0.751 & 1.340 & 0.514 \\
w/ GARL (subset A) & \textbf{4.218} & \textbf{0.652} & \textbf{1.124} & \textbf{0.568} \\ 
&  \textcolor{red}{+20.89\%} & \textcolor{red}{+13.18\%}& \textcolor{red}{+16.12\%}& \textcolor{red}{+10.51\%}\\
\bottomrule
\end{tabular}
}
\label{tab:ablation-result-sgl}
\end{table}

\begin{table}[t]
\centering
\caption{Ablation about Region Aware Feature Training (\textbf{RAFT}).}
\resizebox{0.48\textwidth}{!}{
\begin{tabular}{@{}ccccc@{}}
\toprule
Approach Setting & \multicolumn{1}{l}{ADE ($\downarrow$)} & \multicolumn{1}{l}{ALDE ($\downarrow$)} & \multicolumn{1}{l}{APE ($\downarrow$)} & \multicolumn{1}{l}{MnRE ($\uparrow$)} \\ \midrule
w/o RAFT (500 objects) & 9.625 & 0.669 & 0.812 & 0.571 \\
w/ RAFT (500 objects) & \textbf{9.461} & \textbf{0.661} & \textbf{0.767} & \textbf{0.576} \\ 
&  \textcolor{red}{+1.70\%}& \textcolor{red}{+1.20\%}& \textcolor{red}{+5.54\%}& \textcolor{red}{+0.87\%}\\

\midrule
w/o RAFT (subset B) & 6.302 & 0.684 & 1.224 & 0.555 \\
w/ RAFT (subset B) & \textbf{5.602} & \textbf{0.634} & \textbf{1.117} & \textbf{0.583} \\ 
&  \textcolor{red}{+11.17\%}& \textcolor{red}{+7.30\%}& \textcolor{red}{+8.74\%}& \textcolor{red}{+5.06\%}\\
\bottomrule
\end{tabular}
}
\label{tab:ablation-result-raft}
\end{table}

\begin{table}[t]
\centering
\caption{Ablation about \textbf{thickness} and \textbf{Gaussian} based integration}
\resizebox{0.48\textwidth}{!}{
\begin{tabular}{@{}ccccc@{}}
\toprule
Approach Setting & \multicolumn{1}{l}{ADE ($\downarrow$)} & \multicolumn{1}{l}{ALDE ($\downarrow$)} & \multicolumn{1}{l}{APE ($\downarrow$)} & \multicolumn{1}{l}{MnRE ($\uparrow$)} \\ \midrule
THICKNESS (500 objects) & 12.498 & 0.811 & 1.803 & 0.537 \\
GAUSSIAN (500 objects) & \textbf{9.461} & \textbf{0.661} & \textbf{0.767} & \textbf{0.576} \\ 
&  \textcolor{red}{+24.30\%}& \textcolor{red}{+18.50\%}& \textcolor{red}{+57.46\%}& \textcolor{red}{+7.26\%}\\
\bottomrule
\end{tabular}
}
\label{tab:ablation-result-thickness}
\end{table}

\subsection{Application}

We apply the reconstructed results with physical properties from PUGS to object grasping tasks. Firstly, we collect videos of the objects from real-world scenarios using a smartphone. These videos are then processed to extract frames and get multi-view images. Then we use COLMAP to estimate the camera parameters. Based on the calibrated multi-view images, PUGS are employed to reconstruct these objects and predict their Young’s modulus, allowing us to assess the deformability of the objects. Since NeRF2Physics\cite{zhai2024physical} can also predict various physical properties, we conducted a comparative analysis with the results produced by it.

As shown in the Fig.~\ref{fig:teaser}, we conduct an experiment on a cotton package. The opening size of robotic gripper can be adjusted based on the predicted Young’s modulus. A higher Young’s modulus indicates lower deformability, requiring a larger opening size. NeRF2Physics\cite{zhai2024physical} incorrectly predicted the Young’s modulus of the object (30+ GPa), resulting in an opening size nearly equal to the object’s width, which ultimately leads to a failed grasp. In contrast, PUGS accurately predicted the physical property (0.5+ GPa), enabling the gripper to successfully grasp the object.

\section{Conclusion}

In this paper, we propose PUGS, a 3DGS-based framework capable of densely reconstructing for object's physical properties from multi-view images. Experimental results and visualizations demonstrate that PUGS achieves accurate and coherent predictions of material and physical properties, while also benefiting downstream tasks in robotics. Future work includes expanding the framework to enable physical understanding at the scene level.

\newpage
\bibliographystyle{IEEEtran}
\balance
\bibliography{ref}
\end{document}